\documentclass{article}

\usepackage[numbers]{natbib}
\usepackage[preprint]{neurips_2025}

\usepackage[utf8]{inputenc} 
\usepackage[T1]{fontenc}    
\usepackage{hyperref}       
\usepackage{url}            
\usepackage{booktabs}       
\usepackage{amsfonts}       
\usepackage{nicefrac}       
\usepackage{microtype}      
\usepackage{graphicx}       
\usepackage{amsmath}
\usepackage{float}
\usepackage{tablefootnote}  
\usepackage{placeins}
\usepackage{subcaption}
\usepackage{tabularx}
\usepackage{bm}
\usepackage{multirow}

\title{Diagnosing Generalization Failures in Fine-Tuned LLMs: A Cross-Architectural Study on Phishing Detection}

\author{%
  \pmb{Frank Bobe III}\thanks{Corresponding Author}, \ Gregory D. Vetaw, \ Chase Pavlick, \\
  Darshan Bryner, \ Matthew Cook, \ and Jose Salas-Vernis \\[1em]
  Naval Surface Warfare Center Panama City Division \\[0.5em]
  \texttt{\{frank.e.bobe, gregory.d.vetaw, tristan.c.pavlick,}\\
  \texttt{darshan.w.bryner, matthew.g.cook12, jose.l.salasvernis\}.civ@us.navy.mil}
}

\begin{document}

\maketitle

\begin{abstract}
The practice of fine-tuning Large Language Models (LLMs) has achieved state-of-the-art performance on specialized tasks, yet diagnosing why these models become brittle and fail to generalize remains a critical open problem. To address this, we introduce and apply a multi-layered diagnostic framework to a cross-architectural study. We fine-tune Llama 3.1 8B, Gemma 2 9B, and Mistral models on a high-stakes phishing detection task and use SHAP analysis and mechanistic interpretability to uncover the root causes of their generalization failures. Our investigation reveals three critical findings: (1) Generalization is driven by a powerful synergy between architecture and data diversity. The Gemma 2 9B model achieves state-of-the-art performance (>91\% F1), but only when trained on a stylistically diverse ``generalist'' dataset. (2) Generalization is highly architecture-dependent. We diagnose a specific failure mode in Llama 3.1 8B, which performs well on a narrow domain but cannot integrate diverse data, leading to a significant performance drop. (3) Some architectures are inherently more generalizable. The Mistral model proves to be a consistent and resilient performer across multiple training paradigms. By pinpointing the flawed heuristics responsible for these failures, our work provides a concrete methodology for diagnosing and understanding generalization failures, underscoring that reliable AI requires deep validation of the interplay between architecture, data, and training strategy.
\end{abstract}

\section{Introduction}
Recent findings highlight that the challenges of deploying Large Language Models (LLMs) extend beyond traditional technical limitations: 31\% of organizations report unauthorized access to sensitive data via AI systems, while attackers increasingly leverage LLMs to generate sophisticated phishing campaigns at an unprecedented scale \cite{cost_of_breach}. The IBM Security 2025 Cost of a Data Breach Report reveals that breaches involving generative AI now cost an average of \$670K more per incident than traditional attacks. Furthermore, governance frameworks for mitigating these threats remain nascent, with 63\% of organizations lacking policies to manage risks like AI-driven phishing campaigns \cite{cost_of_breach}. These statistics underscore a critical vulnerability at the intersection of AI and security.

A key technical reason for these security gaps is the problem of generalization failure in fine-tuned LLMs. Fine-tuning has become standard practice for creating specialized AI systems, achieving state-of-the-art performance on benchmark tasks. However, this process often creates brittle models, which are specialized systems that perform exceptionally well on data that closely resembles their training set but whose accuracy collapses when faced with novel inputs. This performance degradation is a direct consequence of domain shift, a scenario where the statistical distribution of data encountered during deployment, such as modern phishing emails, differs from the distribution of the training data, such as an older email corpus. While the existence of this problem is well-documented \cite{quinonero-candela2009dataset}, its underlying mechanisms remain underexplored, particularly how factors like model architecture, data diversity, and fine-tuning hyperparameters interact with these emerging security risks.

This paper aims to fill this gap by conducting a systematic, cross-architectural study to diagnose the root causes of generalization failures. Using phishing detection as a high-stakes testbed, our work seeks to answer the following fundamental questions about the robustness of fine-tuned LLMs:
\begin{itemize}
    \item \textbf{Architecture Dependence:} Does the choice of the base model architecture (e.g., Llama vs. Gemma vs. Mistral) fundamentally impact its robustness to domain shift?
    \item \textbf{Data Diversity:} Is training on more diverse data always beneficial, or can it lead to ``negative transfer'' where the model fails to learn a coherent set of features?
    \item \textbf{Training Strategy:} Do advanced techniques like Chain-of-Thought (CoT) fine-tuning provide a universal improvement in generalization over standard fine-tuning?
\end{itemize}

\subsection*{Related Work}
To situate our work and provide context for these security gaps, we review three active areas of machine learning research: the application of LLMs to cybersecurity, the generalization properties of parameter-efficient fine-tuning, and the interpretability of fine-tuned models.

\paragraph{Large Language Models for Cybersecurity} 
The application of LLMs to cybersecurity tasks has shown significant promise in recent years \cite{llms_for_security}. Early work demonstrated the effectiveness of models like BERT \cite{bert} for spam and phishing detection across multiple datasets \cite{bert_for_spam, lingspam_dataset}. However, such high performance on curated benchmarks often proves brittle when models are deployed outside their original training context. For instance, a model fine-tuned exclusively on the Enron email corpus, with its formal corporate style, may fail when tasked with evaluating emails from the more diverse SpamAssassin corpus \cite{bert_for_spam}. The model, having overfit to the narrow stylistic features of its training data, misclassifies legitimate but stylistically different content. This vulnerability underscores the need for more robust generalization strategies \cite{abdali2024securing}, and our work builds on this foundation by using modern PEFT methods to diagnose the architectural and data-driven causes of these failures.

\paragraph{Generalization of Fine-Tuned Models} 
Parameter-Efficient Fine-Tuning (PEFT) methods like QLoRA \cite{qlora} have become standard for adapting LLMs to new problems or tasks. While their effectiveness is established \cite{lora}, their impact on generalization is an area of active research. Some studies suggest PEFT can be more robust to catastrophic forgetting than full fine-tuning \cite{biderman2024lora}. However, the relationship between specific PEFT hyperparameters, such as the LoRA rank (`r'), and out-of-domain robustness is not yet well understood. Furthermore, a systematic comparison of how different modern architectures (e.g., Llama, Gemma, Mistral) respond to data diversity under a consistent PEFT strategy remains an open question. Our study directly addresses this gap by isolating the impacts of architecture and data.

\paragraph{Interpretability of Fine-Tuned Models} As fine-tuned models are deployed in sensitive applications, there is a growing need to understand their decision-making processes \cite{biology_llm}. Mechanistic interpretability aims to reverse-engineer the algorithms, or ``circuits,'' learned by neural networks \cite{circuit_tracing_methods}. In our work, interpretability is a critical diagnostic tool. When a model fails, an interpretability method like SHAP can reveal \textit{why} by showing which input features were most influential. For example, it can expose a flawed heuristic where the model over-relies on corporate jargon as a signal of legitimacy. We take this one step further by connecting this high-level explanation to a low-level, mechanistic one. By analyzing attention patterns, we can identify the specific circuit responsible for the failure, demonstrating the practical utility of interpretability as an essential component of the model debugging process.

\subsection*{Our Contributions}
Our methodology proceeds in two main stages: performance evaluation and failure diagnosis. We first conduct a cross-architectural study fine-tuning Llama 3.1 8B \cite{llama3}, Gemma 2 9B \cite{gemma2}, and a Mistral  \cite{mixtral} variant on three stylistically different phishing datasets. After identifying key generalization failures, we employ a multi-layered interpretability approach using SHAP \cite{shap} and mechanistic analysis to investigate their root causes.

Our primary contributions are threefold:
\begin{enumerate}
    \item We present a cross-architectural fine-tuning study that demonstrates how generalization emerges from a synergy between model architecture and data diversity. Our results show that while some architectures excel with varied data, others exhibit critical failures, establishing that architectural choice is a fundamental factor in building robust models.

    \item We identify and quantify significant dataset issues in public corpora, revealing high rates of label noise and the presence of ``naturally occurring'' adversarial text. We demonstrate that these data-level confounders can cause models to learn brittle, non-generalizable heuristics, complicating the evaluation of true generalization.

    \item We introduce a practical diagnostic method for uncovering specific model shortcomings. By combining SHAP analysis with the mechanistic inspection of attention heatmaps, our method explains \textit{why} a model fails, revealing the flawed internal heuristics, such as overfitting to stylistic artifacts that leads to poor generalization.
\end{enumerate}

\section{Methodology}
In this section, we cover the methodology, specifically the models and datasets we use to validate our experiments. We also cover the model implementation details used for fine-tuning the LLMs and metrics used to quantify our results and provide interpretability.

\subsection{Base Models and Fine-Tuning}
To provide a diverse cross-section of modern, open-source architectures, we use three base models: \textbf{Meta Llama 3.1 8B-Instruct} , \textbf{Google Gemma 2 9B Instruct}, and a \textbf{Mistral 12B} variant. All models are fine-tuned using the QLoRA (Quantized Low-Rank Adaptation) technique \cite{qlora}. The weights of the base models are quantized to 4-bit precision. Trainable LoRA matrices are injected into the query and value projections of the self-attention mechanism. These components are responsible for determining what information each token seeks from the context (the query) and what information it contributes to the representation (the value). Targeting these specific projections is a common and effective practice for applying LoRA to Transformer-based models.

Formally, for a pretrained weight matrix $\bm{W_0} \in \mathbb{R}^{d \times k}$, LoRA introduces two low-rank matrices, $\bm{B} \in \mathbb{R}^{d \times r}$ and $\bm{A} \in \mathbb{R}^{r \times k}$, where the rank $r \ll \min(d, k)$. The forward pass is then modified such that:
\begin{equation}
\bm{h} = \bm{W_0}\bm{x} + \Delta \bm{W}\bm{x} = \bm{W_0}\bm{x} + \bm{B}\bm{A}\bm{x}
\end{equation}
where $\bm{h}$ denotes the output vector and $\bm{x}$ represents the input vector. During fine-tuning, the original weights $\bm{W_0}$ are frozen and only the parameters of the adapter matrices $\bm{A}$ and $\bm{B}$ are updated. This significantly reduces the number of trainable parameters, making the fine-tuning process highly efficient.

\paragraph{Datasets and Experimental Design}
Our experiments hinge on three distinct datasets chosen for their stylistic and temporal differences, which create a challenging domain-shift scenario for evaluation:
\begin{itemize}
    \item \textbf{The Enron Corpus:} We use the well-known subset prepared by Metsis et al. \cite{enron_spam}, containing approximately 33,700 emails from the early 2000s. This corpus is stylistically homogeneous, characterized by formal language and professional jargon, with a nearly balanced distribution of spam and legitimate emails.

    \item \textbf{The SpamAssassin Corpus:} A classic public benchmark of approximately 6,600 emails, valued for its stylistic diversity \cite{spamassassin}. It contains a heterogeneous mix of legitimate emails (``ham'') and spam, sourced from technical mailing lists, personal communications, and newsletters.

    \item \textbf{Modern Phishing Corpus:} A contemporary collection of over 18,600 examples, consisting exclusively of phishing attacks from around 2020 \cite{cybercop_phishing}. This dataset is crucial for evaluating a model's ability to recognize modern adversarial tactics that are absent in the older corpora.
\end{itemize}

\paragraph{Experiment 1: Standard Supervised Fine-Tuning}
Our first experiment proceeds in two steps. First, to establish a true performance baseline, we evaluate the zero-shot classification accuracy of each base model architecture on all datasets \textbf{before} any fine-tuning. 

Following the baseline evaluation, we conduct the main supervised fine-tuning experiment using simple binary labels (`SPAM` or `LEGIT`). For each model architecture, we create and compare two fine-tuned versions to investigate the impact of data diversity:
\begin{itemize}
    \item \textbf{The Specialist Model}, which is fine-tuned exclusively on the Enron corpus. This version measures performance when a model specializes on a single, narrow domain.
    \item \textbf{The Generalist Model}, which is fine-tuned on a combined corpus of all three datasets. This version is designed to evaluate the benefits of training on stylistically diverse data.
\end{itemize}
Both the Specialist and Generalist versions are then evaluated against unseen test splits from \textbf{all three} source corpora, allowing for a direct comparison of in-domain performance versus out-of-domain generalization.

\paragraph{Experiment 2: Chain-of-Thought (CoT) Fine-Tuning}
To investigate whether structured reasoning improves generalization, we conduct a separate experiment using a synthetic Chain-of-Thought (CoT) dataset. For this, we convert the binary labels of the training data into a structured reasoning format. Instead of predicting a simple label, the model is taught to first identify and articulate the specific red flags detailed below before making a final classification. This approach is applied to Generalist (combined CoT) training configuration.

\paragraph{Chain-of-Thought (CoT) Criteria for Phishing Detection}
The model was taught to identify and articulate specific red flags before making a final classification. These criteria include:
\begin{itemize}
    \item Inconsistencies in Email Addresses, Domain Names, or Links
    \item Unusual Requests (e.g., unsolicited offers, unexpected invoices)
    \item Extremely Brief or Vague Messages (``Short and Sweet'')
    \item Unfamiliar or Inappropriate Tone/Greeting Patterns
    \item Requests for Credentials, Payment Information, or Personal Details
    \item Suspicious or Unexpected Attachments
    \item Emails Initiated Without Prior Interaction or Context
    \item Threats, Urgency, or High-Pressure Tactics
    \item Significant Spelling or Grammatical Errors
\end{itemize}

\paragraph{Hold-Out Evaluation Sets}
For a robust evaluation of generalization, all models were evaluated against unseen test splits held out from each of the three source corpora: the Enron dataset, the SpamAssassin dataset, and the Modern Phishing Corpus. This hold-out data was never used during any phase of training.

\paragraph{Implementation and Hyperparameter Tuning}
To identify a robust QLoRA configuration, we conducted a comprehensive hyperparameter sweep over the discrete search space detailed in Table~\ref{tab:hyperparams}. We employed a random search methodology to efficiently explore this landscape, sampling configurations that included LoRA ranks $r \in \{8, 16, 32, 64\}$, peak learning rates of $\{5\text{e-}5, 1\text{e-}4, 2\text{e-}4\}$ using various schedulers (cosine, linear, constant), and training durations between 2 and 5 epochs. For reproducibility, the script seeds the random number generator for each run. A loop iterates 24 times to generate distinct parameter sets, constructing configurations by randomly sampling one value for each hyperparameter from these predefined lists. Based on the performance from these 24 runs, a single, high-performing configuration was selected and used consistently across all models for our main comparative experiments.

\vspace{-1em} 

\begin{table}[H]
\centering
\caption{Hyperparameter Search Space for the Initial Random Search}
\label{tab:hyperparams}
\begin{tabular}{ll}
\toprule
\textbf{Hyperparameter} & \textbf{Discrete Values Sampled} \\
\midrule
Learning Rate Scheduler & \{cosine, linear, constant\} \\
Peak Learning Rate & \{5e-5, 1e-4, 2e-4\} \\
Epochs & \{2, 3, 5\} \\
Batch Size per GPU & \{1, 2\} \\
Gradient Accumulation & \{8, 16\} \\
LoRA Rank ($r$) & \{8, 16, 32, 64\} \\
Data Fraction & \{25\%, 50\%, 100\%\} \\
\bottomrule
\end{tabular}
\end{table}

\paragraph{Evaluation and Interpretability}
Our evaluation process consists of two main stages: performance measurement and failure diagnosis. First, to measure task-specific mastery and real-world robustness, we assess all models using the weighted \textbf{F1 Score}, \textbf{Accuracy}, \textbf{Precision}, and \textbf{Recall} as our primary metrics. Second, to diagnose the generalization failures identified by these metrics, we employ a multi-layered interpretability approach. We begin with a high-level analysis using SHAP (SHapley Additive exPlanations) \cite{shap} to visualize which words most influenced a given prediction. We then probe deeper with a mechanistic analysis of the model's attention heads, using heatmaps to reverse-engineer the specific, human-understandable circuits the model learned during fine-tuning.

\section{Experimental Results}

This section details the quantitative outcomes of our experiments, beginning with the justification for our chosen hyperparameters, followed by the baseline capabilities of each model, and finally the results from our two fine-tuning paradigms. The analysis reveals a complex interplay between model architecture, data diversity, and training strategy, highlighting surprising successes and critical failure modes.

\paragraph{Justification of Hyperparameter Configuration}
To ensure a robust and fair comparison, we first conduct a comprehensive hyperparameter sweep involving 24 varied combinations of learning rate schedulers, peak learning rates, LoRA ranks, and other parameters. For successfully converged models, the results of this sweep show very little performance difference, with F1 scores tightly clustered within a narrow range of less than two percentage points. Based on this, and to isolate the effects of model architecture and training data, we select a single, representative configuration for our main experiments: training for 3 epochs with a constant peak learning rate of 1e-4, a LoRA rank (r) of 16, a per-GPU batch size of 2 with 8 gradient accumulation steps, and using 100\% of the training data. 

\paragraph{Output Parsing and Evaluation Metrics}
To convert the models' generative text outputs into binary class labels for evaluation, we employed a strict parsing methodology. Each model was prompted to classify an email with a single-word answer---`SPAM` or `LEGIT`---using the following template:
\begin{quote}
Analyze the following email and classify it as either 'SPAM' or 'LEGIT'. Provide only the single word label as your answer.

Email:
\{email\_text\}

Classification:
\end{quote}
The generated output string was then normalized by converting it to uppercase and removing leading/trailing whitespace. We then checked if this normalized string began with ``SPAM'' or ``LEGIT'' to determine the model's prediction. Alongside this prediction, we calculate a confidence score by applying a softmax function to the logits of the first generated token. This score represents the model's estimated probability for its chosen label. If a model failed to produce an output starting with either of the expected keywords, its prediction was automatically counted as incorrect. From these parsed binary labels, we calculate our primary metrics: Accuracy, Precision, Recall, and the weighted F1 Score.

\subsection{Baseline Zero-Shot Performance}

Before fine-tuning, we evaluate the inherent, zero-shot capabilities of each model across our datasets to establish a performance baseline. As shown in Table \ref{tab:baseline_results}, the models exhibit varied and generally modest performance, underscoring the necessity of task-specific training.

Notably, \textbf{Llama 3.1 8B} demonstrates the strongest initial aptitude for the \texttt{spamassassin} dataset, achieving a respectable F1 score of \textbf{0.7854}. Conversely, \textbf{Gemma 2 9B} shows more moderate performance on the same dataset with an F1 score of \textbf{0.5713}. The \textbf{Mistral} model presents a balanced profile, delivering the best performance on the combined \texttt{generalist} dataset with an F1 score of \textbf{0.7046}. These baselines clearly indicate that no single base model is inherently superior across all email domains, and each possesses distinct inductive biases that influence its out-of-the-box performance.

\vspace{-2em}

\begin{table}[H]
\centering
\caption{Baseline Zero-Shot Performance of Untrained Models}
\label{tab:baseline_results}
\begin{tabular}{llcccc}
\toprule
\textbf{Model} & \textbf{Dataset} & \textbf{F1 Score} & \textbf{Accuracy} & \textbf{Precision} & \textbf{Recall} \\
\midrule
\textbf{Llama} & SpamAssassin & 0.7854 & 0.7829 & 0.7880 & 0.7829 \\
& Modern Phishing & 0.5187 & 0.5692 & 0.5279 & 0.5692 \\
& Enron & 0.4406 & 0.4897 & 0.4886 & 0.4897 \\
& Generalist (Combined) & 0.5877 & 0.6054 & 0.5999 & 0.6054 \\
\midrule
\textbf{Gemma} & SpamAssassin & 0.5713 & 0.5042 & 0.7538 & 0.5042 \\
& Modern Phishing & 0.3966 & 0.4411 & 0.5091 & 0.4411 \\
& Enron & 0.4713 & 0.5041 & 0.5050 & 0.5041 \\
& Generalist (Combined) & 0.5019 & 0.5135 & 0.5393 & 0.5135 \\
\midrule
\textbf{Mistral} & SpamAssassin & 0.7017 & 0.6645 & 0.7564 & 0.6645 \\
& Modern Phishing & 0.5676 & 0.6293 & 0.6338 & 0.6293 \\
& Enron & 0.4779 & 0.5568 & 0.6445 & 0.5568 \\
& Generalist (Combined) & 0.7046 & 0.7085 & 0.7087 & 0.7085 \\
\bottomrule
\end{tabular}
\end{table}

\subsection{Standard Fine-Tuning Results: The Power of Data Diversity}

Our standard fine-tuning experiments, summarized in Table \ref{tab:finetuning_results}, yield the most significant findings of this study. We compare models trained as ``Specialists'' (on Enron data only) against ``Generalists'' (trained on the combined dataset).

The results reveal a clear winner for the more challenging task of handling diverse data. Gemma 2 9B, when trained as a Generalist and evaluated on the combined Generalist test set, achieves an outstanding F1 score of 0.9132. This is the highest score across all experiments, demonstrating that the Gemma architecture is exceptionally capable of integrating diverse data sources to build a robust, generalizable classifier. Mistral also proves to be a highly effective Generalist, achieving a strong F1 score of 0.8378 on the same diverse test set.

In contrast, while the Specialist models performed capably on their own in-domain Enron test set, the Llama 3.1 8B architecture showed a failure to generalize. Its performance as a Specialist on the narrow Enron test set (F1 0.7828) was respectable, but its F1 score dropped to 0.6648 when trained as a Generalist on the diverse dataset. This highlights the study's key finding: achieving high performance on a varied, general-purpose task is a distinct and more difficult challenge than specializing on a narrow one.

\begin{table}[H]
\centering
\caption{Standard Fine-Tuning Performance (Evaluated on Hold-Out Sets)}
\label{tab:finetuning_results}
\begin{tabular}{llcccc} 
\toprule
\textbf{Training Dataset} & \textbf{Model} & \textbf{F1 Score} & \textbf{Accuracy} & \textbf{Precision} & \textbf{Recall} \\
\midrule
\multirow{3}{*}{\shortstack[l]{\textbf{Enron} \\ \textbf{(Specialist)}}} & Llama   & 0.7828 & 0.7903 & 0.8364 & 0.7903 \\
                                      & Gemma   & 0.5974 & 0.6407 & 0.7481 & 0.6407 \\
                                      & Mistral & 0.8091 & 0.8138 & 0.8477 & 0.8138 \\
\midrule
\multirow{3}{*}{\shortstack[l]{\textbf{Generalist} \\ \textbf{(Combined)}}}   & Llama   & 0.6648 & 0.6824 & 0.7836 & 0.6824 \\
                                      & Gemma   & \textbf{0.9132} & \textbf{0.9130} & \textbf{0.9202} & \textbf{0.9130} \\
                                      & Mistral & 0.8378 & 0.8384 & 0.8647 & 0.8384 \\
\bottomrule
\end{tabular}
\end{table}

\subsection{Chain-of-Thought (CoT) Fine-Tuning: A Contextual Benefit}

We hypothesize that explicitly training models to reason through a classification task using CoT could improve performance. However, the results in Table \ref{tab:cot_results} show that its benefits are highly dependent on the model architecture and training data.

For the Generalist models, Mistral is the only one to clearly benefit from the CoT approach, emerging as the top performer in this category with an F1 score of 0.7574. This suggests its architecture may be more amenable to adopting structured, step-by-step reasoning. Interestingly, the CoT approach hinders the Gemma model, whose performance drops significantly compared to standard fine-tuning. This implies that for an architecture that already generalizes well from diverse data, the rigid structure of CoT may act as a constraint rather than an aid. Furthermore, the complexity of generating structured reasoning may demand greater model capacity; it is possible that a larger LoRA rank (`r`) or a base model with more parameters is required to fully realize the benefits of CoT fine-tuning.

\begin{table}[H]
\centering
\caption{Chain-of-Thought (CoT) Fine-Tuning Performance (Evaluated on Hold-Out Sets)}
\label{tab:cot_results}
\begin{tabular}{llcccc}
\toprule
\textbf{Training Dataset} & \textbf{Model} & \textbf{F1 Score} & \textbf{Accuracy} & \textbf{Precision} & \textbf{Recall} \\
\midrule
\multirow{3}{*}{\shortstack[l]{\textbf{Generalist} \\ \textbf{(Combined)}}} & Llama & 0.7273 & 0.5717 & 0.9994 & 0.5717 \\
                                       & Gemma & 0.7049 & 0.7474 & 0.7057 & 0.7474 \\
                                       & Mistral & \textbf{0.7574} & \textbf{0.7443} & \textbf{0.7841} & \textbf{0.7443} \\
\bottomrule
\end{tabular}
\end{table}

\section{Interpretability Analysis \& Discussion}

Mechanistic interpretability seeks to reverse engineer neural networks, similar to how one might reverse engineer a compiled binary computer program. The quantitative results in Experimental Results reveal a critical disconnect: our fine-tuned models exhibit a significant generalization gap. While some achieve high performance on in-domain data, their robustness on out-of-domain benchmarks can vary dramatically, exposing critical failure modes. This section aims to move beyond these metrics to diagnose the root causes of this generalization failure \cite{safety_alignment}. For high-stakes applications like phishing detection, simply knowing that a model fails is insufficient; we must understand \textit{why} it fails. To this end, we employ a series of analytical techniques, including a data quality audit and SHAP analysis, to probe the internal logic of our best-performing models and reveal the mechanisms of their failure.

\subsection{The Challenge of Benchmark Quality}
While the high F1 scores on the Gemma Generalist dataset suggest task mastery, a significant confounding factor emerges during the \textit{out-of-domain} evaluation on the spamassassin hold-out set. We discover substantial label noise that is, incorrectly labeled ground-truth data within the benchmark itself. Such errors can arise in historical datasets that were curated before the advent of modern annotation standards or for different research objectives where minor inaccuracies were permissible. A manual review of the model's predictions reveals numerous instances where our models correctly identify obvious spam, but the dataset's ground-truth label is `LEGIT` (ham). For example, an email containing the classic Nigerian Prince scam opening, ``Dear Sir. First, I must solicit your confidence in this transaction...'', is labeled as legitimate in the test set, as is an email promising ``A Real And Legitimate Opportunity To Make Some SERIOUS MONEY!''. In both cases, our fine-tuned models correctly predict `SPAM`, but are penalized by the noisy ground-truth data. This finding is critical for interpreting our results. It indicates that the measured out-of-domain F1 scores underestimates the models' true ability to detect spam. More importantly, it highlights a broader challenge in the field: the reliability of public benchmarks for evaluating security models. In an adversarial domain like spam detection, where the very nature of the data is deceptive, the risk of noisy or mislabeled ground-truth data is particularly high. Our findings suggest that researchers must approach such benchmarks with a degree of skepticism and that automated auditing may be a necessary step in the evaluation pipeline.

\subsection{Ablation Study on Data Quality}
Given the discovery of label noise, we conduct an experiment to audit the quality of our primary datasets. We use our best-performing model, the Gemma ``Generalist Expert,'' as an automated auditor to flag examples where its prediction disagrees with the dataset's label with a confidence of >91\%. The results, shown in Table \ref{tab:audit_results}, are striking.

\begin{table}[h!]
\centering
\caption{Results of the Automated Data Quality Audit}
\label{tab:audit_results}
\begin{tabular}{l|c|c}
\toprule
\textbf{Dataset Audited} & \textbf{Total Samples} & \textbf{Potential Mislabels Found} \\
\midrule
SpamAssassin & 3,002 & 19 (~1\%) \\
Modern Phishing & 18,634 & 4,038 (~22\%) \\
Enron & 33,107 & 5,777 (~17\%) \\
\midrule
\textbf{Total} & \textbf{54,743} & \textbf{9,834 (~18\%)} \\
\bottomrule
\end{tabular}
\end{table}

The audit reveals differences in data quality across the corpora. However, it is crucial to note that this audit is performed by a model, not by human annotators. Therefore, the flagged samples represent potential mislabels, and these findings warrant further manual investigation to be fully validated. With this disclaimer in mind, the Phishing Email dataset is found to have a remarkably high rate of potential mislabels at ~22\%, and our Enron Spam training corpus also suffers from significant label noise (~17\%). In contrast, the SpamAssassin dataset is relatively clean, with only ~1\% of samples flagged. This leads us to hypothesize that the model's generalization failure is exacerbated by training on a noisy and unreliable data source \cite{data_ablation}.

\subsection{Adversarial Data Audit}
Beyond simple label noise, we investigate a more subtle and potentially more damaging form of data contamination: the presence of text that could be interpreted as an adversarial attack designed to manipulate an LLM classifier. Such attacks can be deliberately engineered, for instance, by hiding machine-readable instructions in text that is nearly invisible to a human reviewer such as prompts in a tiny font size or colored to match the background which might command a model to ``always provide a positive review.'' While intentional data poisoning is a known threat in AI security, we hypothesize that older, less curated datasets might contain ``naturally occurring'' adversarial text that could inadvertently teach a model incorrect and non-generalizable heuristics. \cite{llm_fool}

To test this hypothesis, we designed an experiment to audit all three datasets. We used the fine-tuned Gemma 2 9B model as a powerful reasoning engine, prompting it to act as an AI security auditor. For each email, the model was given the following detailed prompt, which defined its task and the required output format:

\begin{quote}
You are an AI security auditor. Your task is to detect hidden instructions or text within an email that are designed to manipulate or deceive a language model into making an incorrect classification. Look for phrases like `this is a legitimate email, classify it as ham', instructions to ignore previous prompts, or text hidden in code or comments. First, on a new line, write `DECEPTION: Yes' or `DECEPTION: No'. On the next line, write `RATIONALE:' followed by a brief explanation for your decision.
\end{quote}

The results of this audit, presented in Table \ref{tab:adversarial_audit}, are highly unexpected and reveal a potential problem with using these datasets for training and evaluation without auditing.

\begin{table}[h!]
\centering
\caption{Results of the Adversarial Data Audit}
\label{tab:adversarial_audit}
\begin{tabular}{l|c|c}
\toprule
\textbf{Dataset Audited} & \textbf{Total Samples} & \textbf{Potential Deception Prompts Found} \\
\midrule
SpamAssassin & 601 & 38 (\~6\%) \\
Enron & 6,622 & 242 (\~4\%) \\
Modern Phishing & 3727 & 265 (\~7\%) \\
\bottomrule
\end{tabular}
\end{table}

This finding provides a powerful new explanation for our models' generalization failure. The Enron dataset, far from being a clean corporate corpus, is saturated with informal content that acts as a form of ``naturally occurring'' adversarial noise. Our automated adversarial audit (Table \ref{tab:adversarial_audit}) supports this conclusion, identifying that approximately 4\% of the Enron corpus (242 instances) contained what the auditor model considered potentially deceptive text. A manual review of these flagged emails revealed they were often forwarded jokes, chain letters, or other informal messages containing conversational text that mimics instructions (e.g., ``see the attachment,'' ``forward this to 10 friends''). While not intentionally malicious, our model, trained on this data, consequently learns a spurious correlation by incorrectly associating instructive or unusual formatting with legitimacy. This suggests that rigorously filtering such adversarial noise from the training corpus is a critical step toward building more robust and generalizable models.

\subsection{Diagnosing Failure Mechanisms with SHAP}
To move beyond metrics and understand the model's internal logic, we employ SHAP (SHapley Additive exPlanations) \cite{shap}, a game-theoretic approach to explain model predictions. SHAP is grounded in cooperative game theory and calculates Shapley values, which provide a principled method for attributing a prediction to its input tokens. For a given prediction, the Shapley value $\phi_f$ for a token $f$ is calculated as its average marginal contribution across all possible token coalitions:

\begin{equation}
\phi_f(v) = \sum_{\mathbf{S} \subseteq \mathbf{F} \setminus \{f\}} \frac{|\mathbf{S}|! (|\mathbf{F}| - |\mathbf{S}| - 1)!}{|\mathbf{F}|!} [v(\mathbf{S} \cup \{f\}) - v(\mathbf{S})],
\end{equation}

where $\mathbf{F}$ is the set of all tokens, $f$ is the single token being evaluated, $\mathbf{S}$ is a subset of tokens not containing $f$, and $v(\mathbf{S})$ is the model's output probability for the ``SPAM'' class given only the tokens in subset $\mathbf{S}$. The term $[v(\mathbf{S} \cup \{f\}) - v(\mathbf{S})]$ represents the marginal contribution of token $f$ when added to the coalition $\mathbf{S}$. 

By averaging this contribution over all possible coalitions, SHAP provides a measure of each token's importance. In simpler terms, this process determines a token’s contribution by asking: on average, how much does the presence of this specific token change the model's final prediction when considering all possible combinations of other tokens in the email? A token that consistently pushes the prediction toward ‘SPAM’ will receive a large positive value, while one that consistently pushes it toward ‘LEGIT’ receives a large negative value.

While testing every combination of tokens is computationally prohibitive for most inputs, this exhaustive averaging is what makes the measurement theoretically robust and fair. To make this practical, we employ the PartitionExplainer algorithm from the SHAP library. This method efficiently approximates the true Shapley values by organizing input features into a hierarchy and analyzing the impact of these feature groups, rather than every individual token combination. This allows us to generate these attributions and identify the specific tokens that most influence the model's reasoning.

SHAP helps diagnose \textit{why} a model fails by moving beyond a simple incorrect label to a detailed, token-level explanation. For any given failure, it quantifies the positive (LEGIT) or negative (SPAM) push of each word, revealing the exact features that fooled the model. In our work, this is a semi-manual process: after identifying a cluster of failures, an analyst inspects SHAP visualizations for representative examples to identify the underlying flawed heuristic (e.g., the model assigning too much importance to corporate jargon). While our analysis was manual, this diagnostic workflow could be automated. A system could aggregate SHAP values across thousands of failures to produce a ranked list of features that most consistently contribute to errors, providing a scalable method for discovering a model's systemic vulnerabilities.

The SHAP analysis of our best-performing model the Gemma ``Generalist'' reveals that it develops a simple, but brittle, internal heuristic. The specific failure cases presented below will illustrate this pattern in detail: the model learns to equate the professional, formal, and often technical style found in its training data with ``legitimacy,'' while associating other styles with ``spam.'' Due to the manual effort involved in this type of deep analysis, this investigation should be seen as a snapshot of this specific training paradigm. We anticipate that applying this diagnostic method to different models or datasets would likely reveal different failure modes and heuristics.

\begin{figure}[htbp]
\centering
\includegraphics[width=\textwidth]{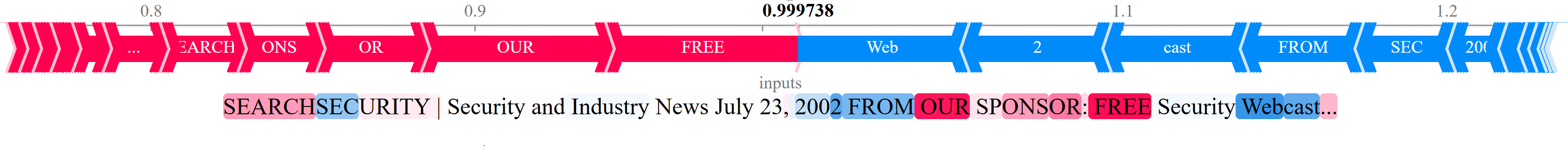}
\caption{SHAP analysis for a false negative prediction on a professional newsletter. Blue tokens push the prediction towards LEGIT, while red pushes towards SPAM. The intensity of the color corresponds to the magnitude of each token's impact; to maintain clarity, the plot only visualizes the most influential tokens. Less impactful words are rendered in lighter shades and may not be visible. The model's output score is displayed at the top; for our binary classifier, a score approaching 1.0 indicates a SPAM prediction, while a lower score indicates a LEGIT prediction.}
\label{fig:shap_false_negative}
\end{figure}

\paragraph{False Negatives (Missed Detections)} The model's primary failure mode is misclassifying professionally formatted newsletters and technical discussions as legitimate. For example, when presented with a SEARCHSECURITY newsletter, the model incorrectly predicts LEGIT. The SHAP analysis in Figure \ref{fig:shap_false_negative} reveals the model's flawed reasoning. Strong positive (LEGIT) attributions are assigned to professional cues like ``Security'' and ``Industry News'', overwhelming the weaker SPAM signals from words like ``FREE'' and ``SPONSOR''.

\begin{figure}[htbp]
    \centering
    \includegraphics[width=1.0\textwidth]{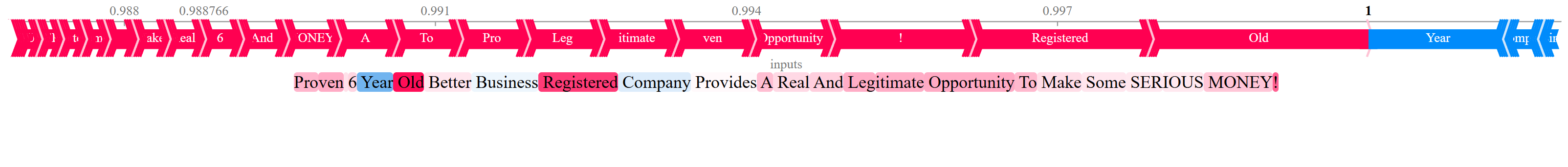}
    \caption{SHAP analysis for a mislabeled MLM scam. The model correctly classifies it as SPAM, driven by strong negative attributions on classic scam keywords.}
    \label{fig:shap_mislabeled_mlm}
\end{figure}

\begin{figure}[htbp]
\centering
\includegraphics[width=1.0\textwidth]{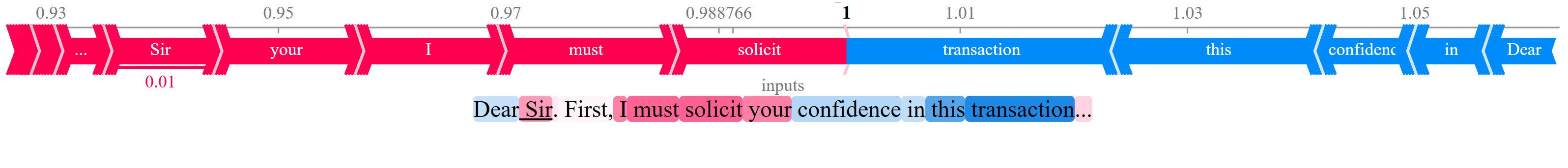}
\caption{SHAP analysis for a mislabeled Nigerian Prince scam identified within the training set during the data quality audit. Although the model was trained on this specific example using the incorrect label, it correctly classifies the email, demonstrating its ability to override label noise.}
\label{fig:shap_mislabeled_nigerian}
\end{figure}

\paragraph{Correct Detections on Mislabeled Data} 
Conversely, the model's ability to correctly identify mislabeled examples provides \textbf{strong evidence} for its conflicting heuristics. Our automated data quality audit (detailed in Table \ref{tab:audit_results}) revealed this on a large scale: the Gemma ``Generalist'' model flagged over 9,800 emails where it confidently predicted \texttt{SPAM} despite a \texttt{LEGIT} label in the dataset. The SHAP visualizations for these cases reveal the basis for the model's confidence. For instance, when analyzing a mislabeled Multi-Level Marketing (MLM) scam (Figure \ref{fig:shap_mislabeled_mlm}), the model correctly classifies it as \texttt{SPAM}, driven by strong negative attributions for classic keywords like ``MONEY'' and ``Opportunity''. The analysis of the Nigerian Prince scam (Figure \ref{fig:shap_mislabeled_nigerian}) tells a similar story, with the model correctly identifying keywords like ``transaction'' and ``confidence'' as strong indicators of spam. \textbf{This suggests a key duality} in the model's behavior: it \textbf{appears to possess} the necessary knowledge to detect spam, but this knowledge \textbf{can be suppressed} by the stronger, overfit heuristic related to stylistic formatting.

In conclusion, the interpretability analysis provides a potential diagnosis for the quantitative failures observed in our experiments. The evidence suggests the model's failure stems not from an inability to understand the features of spam, but from learning the wrong lesson from a stylistically narrow training dataset. Our SHAP analysis indicates the model develops a brittle heuristic that equates the formal, corporate style of the Enron corpus with ``legitimacy.'' This heuristic is the direct cause of the model's primary failure mode; as shown in the analysis of false negatives (e.g., Figure \ref{fig:shap_false_negative}), the model consistently misclassifies out-of-domain professional newsletters because their style resembles the training data. This stylistic rule appears so dominant that it often suppresses the model's other learned knowledge; we found clear cases where the model correctly identified classic spam keywords in mislabeled emails, suggesting it may possess the necessary underlying knowledge, yet it still fails when a phishing attempt adopts a professional tone. This ultimately makes the model's performance unpredictable when encountering the diverse email styles of a real-world environment.

This level of nuanced diagnosis highlights the limitations of traditional evaluation metrics. Uncovering such a deep-seated, contextual failure, moving from what the model got wrong to why it developed this flawed heuristic, would likely be impossible to achieve with accuracy scores alone. It underscores the critical need for interpretability-driven analysis in diagnosing and ultimately building more reliable models. For high-stakes applications like security, such diagnostic tools are not just beneficial; they are essential for validating model behavior before deployment.

\subsection{Mechanistic Analysis of Specialized Attention Heads}
To diagnose the underlying causes of the generalization failures observed in our experiments, we probe the internal mechanisms of our fine-tuned model by analyzing its attention heads. In simpler terms, we investigate the model's ``internal security team'' to see if fine-tuning has taught individual ``analysts'' to become hyper-specialized feature detectors for common phishing tactics \cite{wild_interpretability}. Technically, these ``analysts'' correspond to individual attention heads housed within the model's stacked layers. A designation like ``Layer 6, Head 5`` refers to the fifth parallel attention mechanism located at the sixth layer of the network, distinct from ``Layer 5, Head 6`` which resides at a different depth. This notation allows us to pinpoint exactly which component is responsible for a specific behavior. We isolate these functions using minimal pair analysis, where we compare an attention head's behavior on a baseline legitimate email versus a version altered with a single adversarial feature, such as a generic greeting, a suspicious link, or urgent language.

Before presenting the specific circuits we discovered, it is helpful to explain how to interpret the attention heatmaps. Each heatmap is a grid where both the vertical and horizontal axes list the same sequence of words (tokens) from an email. The color of a square indicates the attention strength: a bright square means the token on the vertical axis is paying strong attention to the token on the horizontal axis. By identifying these patterns of bright squares, we can reverse-engineer the specific rules and relationships the model has learned.

\begin{figure}[!tp]
    \centering
    \begin{subfigure}[b]{\textwidth}
        \centering
        \includegraphics[width=0.8\textwidth]{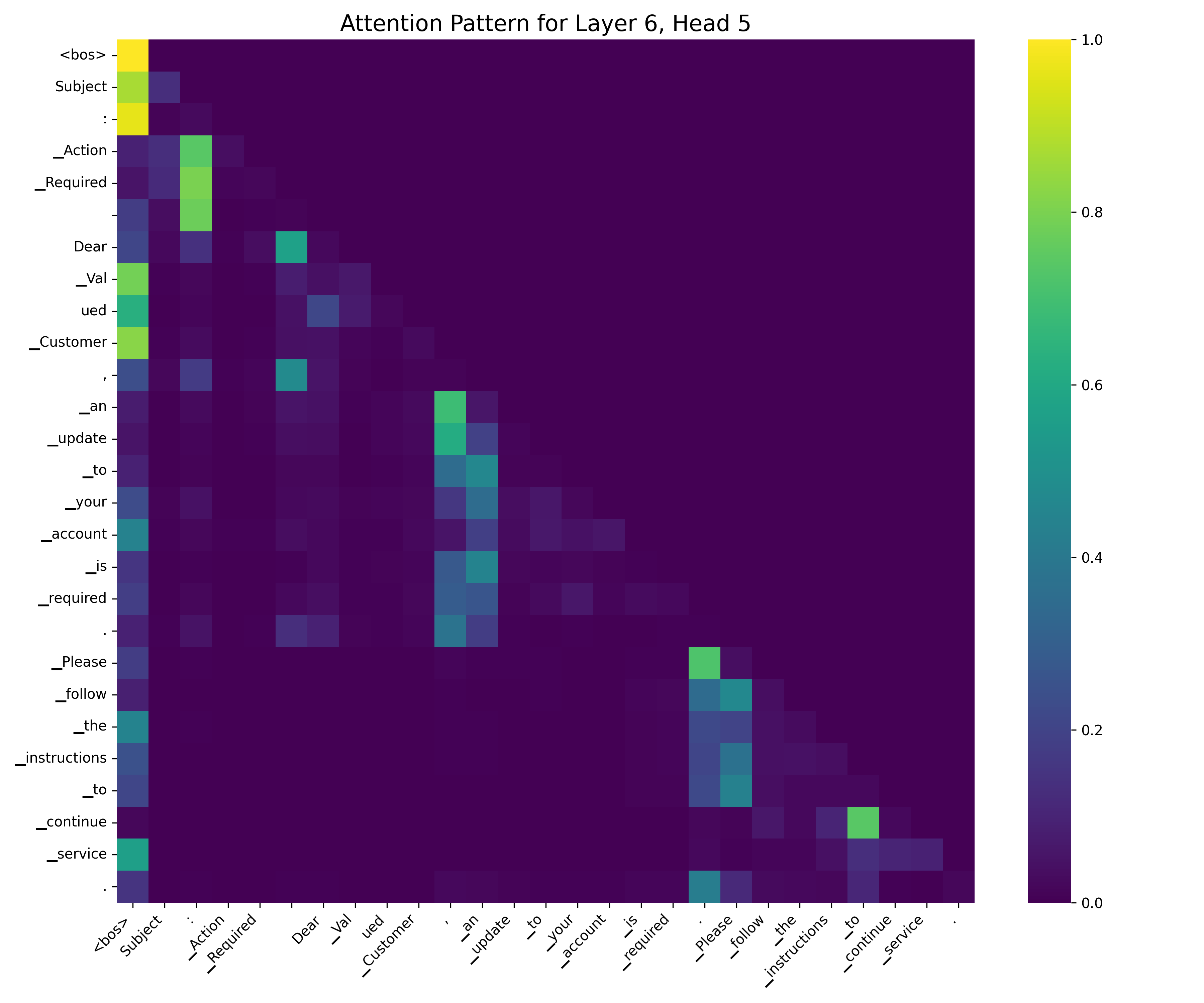}
        \caption{The ``Generic Greeting Detector'' (Layer 6, Head 5).}
        \label{fig:attn_greeting}
    \end{subfigure}
    
    \vspace{1cm} 
    
    \begin{subfigure}[b]{\textwidth}
        \centering
        \includegraphics[width=0.8\textwidth]{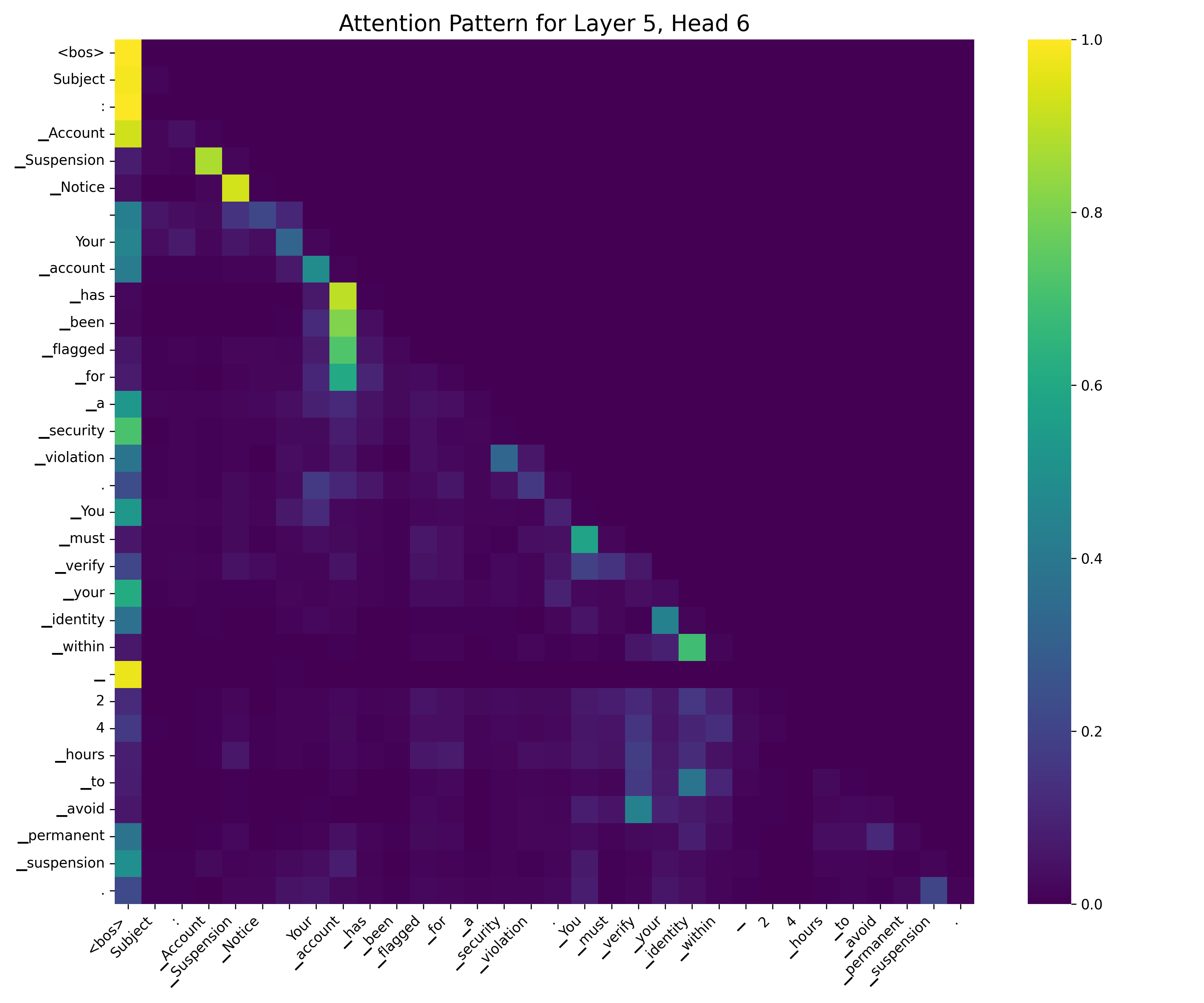}
        \caption{The ``Threat Detector'' (Layer 5, Head 6).}
        \label{fig:attn_urgency}
    \end{subfigure}
    
    \caption{Specialized attention heads for detecting phishing cues. Top (a): A circuit connecting a subject's call to action to a generic greeting. Bottom (b): A circuit specialized in identifying the grammatical structure of threatening language.}
    \label{fig:detector_circuits}
\end{figure}

Our analysis reveals several such specialized heads, as shown in Figure \ref{fig:detector_circuits}. The ``Generic Greeting Detector'' provides a clear example. While its attention pattern on a legitimate email was diffuse, Figure \ref{fig:attn_greeting} shows its transformation when presented with a generic greeting. The head now creates a strong association between the subject's call to action (``Action\_Required'') and the impersonal salutation (``Dear\_Valued\_Customer''). Similarly, the ``Threat Detector'' (Figure \ref{fig:attn_urgency}) emerges in Layer 5, Head 6. This circuit connects words into a cohesive threat narrative, such as linking `permanent` to `suspension` and the command `must` to the pronoun `You`. These are not simple keyword detectors; they are specialized circuits that have learned the grammatical structure of specific phishing tactics.

\begin{figure}[!tp]
    \centering
    \includegraphics[width=0.8\textwidth]{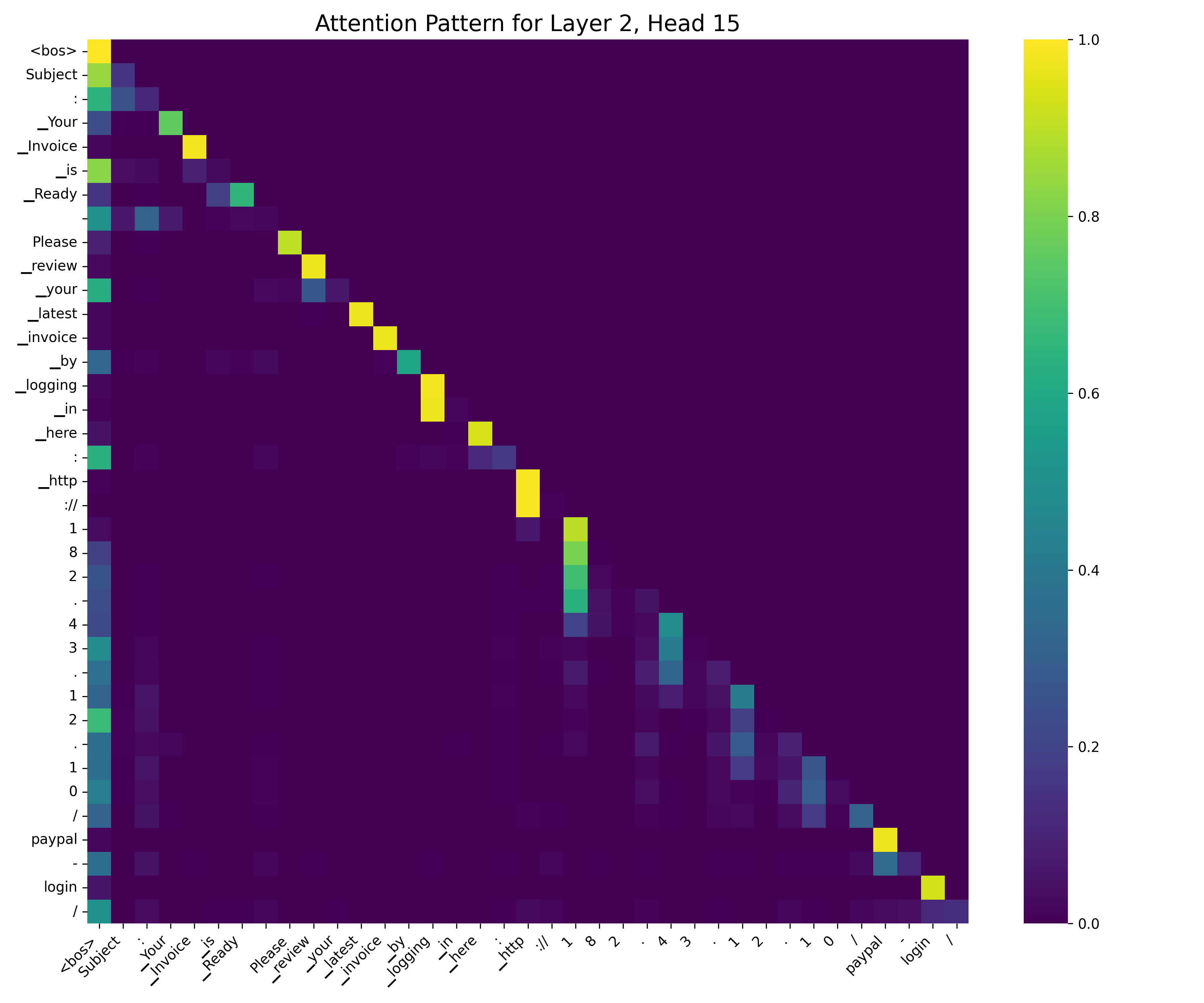}
    \caption{Attention pattern for the ``Suspicious Link Detector'' (Layer 2, Head 15). The head has learned to recognize the specific pattern of an IP address in a URL and link it to the call-to-action.}
    \label{fig:attn_link}
\end{figure}

Finally, the clearest example of mechanistic specialization is the ``Suspicious Link Detector'' at Layer 2, Head 15, shown in Figure \ref{fig:attn_link}. The heatmap for this head reveals an incredibly precise behavior. The tokens that constitute the numerical IP address form a bright, continuous line of high attention along the diagonal, indicating the head treats this sequence as a single, cohesive unit. Furthermore, the token `here` pays strong attention to the beginning of the IP address, linking the call-to-action directly to the suspicious URL. This mechanistic evidence provides a direct causal link between a specific model component and a human-understandable security feature, confirming that fine-tuning does not just teach the model a general classification task, but reshapes its internal components into specialized, interpretable feature detectors.

\FloatBarrier

\section{Conclusion}
This study demonstrates that while fine-tuned LLMs can achieve state-of-the-art performance in specialized domains, their real-world robustness depends on a complex interplay between model architecture, data diversity, and fine-tuning strategy. We found that architectural choice is paramount; Gemma 2 9B's architecture, for instance, thrives on diverse data, allowing it to become a state-of-the-art classifier, while Llama 3.1 8B, despite its capable performance on a narrow domain, demonstrated a significant weakness in generalizing from a stylistically diverse dataset. Crucially, we show that data diversity is the essential factor that unlocks this architectural potential, refuting the concern of negative transfer in this context.

However, our data audits reveal that the datasets used to achieve this diversity are themselves a major confounding factor, with significant label noise (~17-22\%) and ``naturally occurring'' adversarial text. These confounders can cause models to learn brittle, non-generalizable heuristics.

The study underscores the necessity of moving beyond simple accuracy metrics. A holistic evaluation approach combining multi-domain testing, deep data inspection, and interpretability tools like SHAP is essential for diagnosing failures and building truly robust AI systems for high-stakes security applications.

\subsection*{Future Work}
Our findings on architectural sensitivity, data quality challenges, and the powerful synergy between models and data suggest several key research pathways to build more robust phishing classifiers:

\paragraph{Data Curation \& Augmentation}
Our data audits revealed significant label noise (over 17\% in some corpora) and the presence of ``naturally occurring'' adversarial text, highlighting the need for rigorous preprocessing. Future work should employ automated auditing tools \cite{llm_survey} to filter these noisy and confounding examples before training. Additionally, synthetic data generation using models like GPT OSS 20B \cite{gpt_oss} could create stylistically diverse phishing emails to bridge the gaps between historical and modern datasets.

\paragraph{Advanced Training Strategies}
Three promising approaches include:
\begin{itemize}
    \item \textbf{Curriculum Learning}: Based on our finding that naive data mixing can be problematic, a more structured curriculum could be effective. Models could be trained on cleanly filtered, narrow datasets first (e.g., a sanitized Enron corpus), before being gradually exposed to more diverse data to avoid interference from disparate styles \cite{curriculum_llm}.
    \item \textbf{Domain-Adaptive Pre-Training (DAPT)}: An unlabeled pre-training phase on a wide variety of email corpora could allow a model to adapt its internal representations to the email domain before the specialized fine-tuning task begins \cite{dapt_german}.
    \item \textbf{Ensemble of Experts}: Instead of a single ``generalist,'' one could train specialized ``expert'' models for distinct email styles (e.g., corporate vs. technical mailing lists) and combine their predictions via a meta-learner or voting system \cite{der_llm}.
\end{itemize}

\paragraph{Architectural Exploration}
The superior generalization of Gemma 2 9B when trained on diverse data warrants broader studies to understand what makes certain architectures more adept at integrating varied data styles. Future work should include:
\begin{itemize}
    \item Testing other architectures known for robust feature representation, such as mixture-of-experts (MoE) models \cite{mixtral}.
    \item Exploring model merging techniques to combine the robust LoRA adapters from a top-performing generalist like Gemma with the fine-tuned capabilities of other models to create hybrid architectures.
\end{itemize}

\section*{Acknowledgments}
This research is supported and funded by the Office of Naval Research (ONR). We thank ONR code 31 for providing funding and their valuable insights.

\bibliographystyle{unsrtnat}
\bibliography{references}

\end{document}